\pdfoutput=1
\documentclass[11pt]{article}

\usepackage[preprint]{acl}
\usepackage{times}
\usepackage{latexsym}
\usepackage{natbib}

\usepackage[T1]{fontenc}
\usepackage[utf8]{inputenc}

\usepackage{microtype}

\usepackage{inconsolata}

\usepackage{graphicx}

\usepackage{hyperref}       % hyperlinks
\usepackage{url}            % simple URL typesetting
\usepackage{booktabs}       % professional-quality tables
\usepackage{amsfonts}       % blackboard math symbols
\usepackage{nicefrac}       % compact symbols for 1/2, etc.
\usepackage{microtype}      % microtypography
\usepackage{amsthm}
\usepackage{algorithm}
\usepackage{algpseudocode}
\usepackage{mathtools}
\usepackage{graphicx}
\usepackage{subcaption}
\usepackage{tabularx}
\usepackage{dsfont}
\usepackage{multirow}
\usepackage{wrapfig}
\usepackage{array}
\usepackage{amssymb}
\usepackage{multirow}
\usepackage{color}
\usepackage{xcolor}

\usepackage{multicol}

\usepackage{caption}

\usepackage{enumitem}

\usepackage{upgreek}
\usepackage{amsmath,amsfonts,bm}

\def\eqref#1{equation~\ref{#1}}
\def\1{\bm{1}}

\DeclareMathAlphabet{\mathsfit}{\encodingdefault}{\sfdefault}{m}{sl}
\SetMathAlphabet{\mathsfit}{bold}{\encodingdefault}{\sfdefault}{bx}{n}

\theoremstyle{definition}

\definecolor{brightmaroon}{rgb}{0.76, 0.13, 0.28}
\definecolor{darkcyan}{rgb}{0.0, 0.55, 0.55}

\title{Re-ranking Using Large Language Models for Mitigating Exposure to Harmful Content on Social Media Platforms}

\author{
 \textbf{Rajvardhan Oak\textsuperscript{1,3}},
 \textbf{Muhammad Haroon\textsuperscript{1}},
 \textbf{Wonjeong Jo\textsuperscript{1}},\\
 \textbf{Magdalena Wojcieszak\textsuperscript{1}},
 \textbf{Anshuman Chhabra\textsuperscript{2}}
\\
 \textsuperscript{1}University of California, Davis,
 \textsuperscript{2}University of South Florida,
 \textsuperscript{3}Microsoft Corporation, USA
\\
 {
   \texttt{\{rvoak, mharoon, wjo, mwojcieszak\}@ucdavis.edu, anshumanc@usf.edu}
 }
}

\begin{document}
\maketitle
\begin{abstract}
% Social media platforms utilize Machine Learning (ML) and Artificial Intelligence (AI) powered recommendation algorithms to maximize user engagement with online content, which can result in inadvertent exposure to harmful content. While platforms strive to minimize such content, current moderation efforts are limited to classifiers trained on large volumes of human annotated data and/or manual moderation. These approaches are not scalable especially due to the dynamically evolving nature of harm. To rectify this issue of harm classification at scale, we introduce a novel re-ranking approach for harm mitigation via Large Language Models (LLMs) in the zero-shot and few-shot setting. We also propose three novel metrics that can measure whether recommended content is harmful or whether users are recommended non-harmful content aligned with other interests. %aligns with user preferences or exposes them to harmful content. 
% % Through experiments on simulated data from YouTube, we demonstrate the effectiveness of our LLM Re-ranking approach, and empirically show how it can mitigate harm on social media platforms.
Social media platforms utilize Machine Learning (ML) and Artificial Intelligence (AI) powered recommendation algorithms to maximize user engagement, which can result in inadvertent exposure to harmful content.
Current moderation efforts, reliant on classifiers trained with extensive human-annotated data, struggle with scalability and adapting to new forms of harm.
To address these challenges, we propose a novel re-ranking approach using Large Language Models (LLMs) in zero-shot and few-shot settings. Our method dynamically assesses and re-ranks content sequences, effectively mitigating harmful content exposure without requiring extensive labeled data.
Alongside traditional ranking metrics, we also introduce two new metrics to evaluate the effectiveness of re-ranking in reducing exposure to harmful content.
Through experiments on three datasets, three models and across three configurations,  we demonstrate that our LLM-based approach significantly outperforms existing proprietary moderation approaches, offering a scalable and adaptable solution for harm mitigation.

\end{abstract}

\section{Introduction}
\looseness-1 Social media platforms are powered by Machine Learning (ML) and Artificial Intelligence (AI) based recommendation algorithms and models that provide content for users. These algorithms and models are designed to maximize user engagement by learning to recommend content aligned with users’ inferred interests or traits~\cite{covington2016deep}. However, solely optimizing for user engagement metrics can indirectly drive exposure to harmful content. For instance, a teenager interested in fitness may be recommended content promoting eating disorders, users interested in finance may encounter clickbait and scam videos, and a sad adolescent may be directed to content about depression or suicide.% \cite{hilbert20248}. 
As a result, there are serious concerns that platform recommendation systems can indirectly foment misinformation, addictions or mental health crises, and lead to other problems for individuals and society at large~\cite{haidt2022adolescent, haidt2023social, roose2019makingytradicalNYT, tufekci2018youtube}. 

Internally, social media platforms seek to mitigate harmful content using AI/ML classifiers. However, there are two major challenges associated with their use~\cite{gorwa2020algorithmic, chen2021automated, khan2021disinformation}: (1) classifiers often require large volumes of annotated data for training, and (2) categorizing harmful content is a dynamic temporal problem (e.g. a new dangerous challenge for teenagers emerges online). Classifiers cannot automatically generalize to new forms of harm, without having been trained on explicitly labeled data. As a result, harm classifiers are susceptible to concept drift \cite{quinonero2022dataset} and requiring humans to annotate large amounts of data.  

\begin{figure}[t]
\centering
\includegraphics[width=0.48\textwidth]{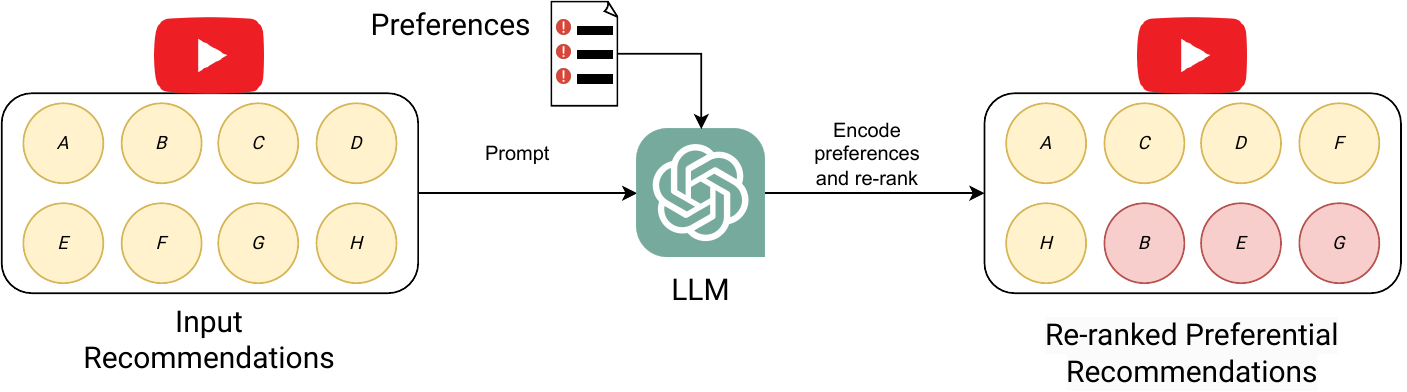}
\caption{An overview of our re-ranking approach for mitigating exposure to harmful content. We prompt the LLM with the input set of recommendations and a set of preference constraints. The LLM re-ranks the the recommendations in accordance with the provided preferences. Here, recommendations \textit{B}, \textit{E}, and \textit{G} are harmful and hence, downranked.}
\label{fig:overview}\vspace{-4mm}
\end{figure}

\looseness-1 In this paper, we propose methods that can circumvent both issues: that of scale as well as that of the dynamicity of harm. We utilize Large Language Models (LLMs) for this purpose as they have been shown to demonstrate stellar reasoning capabilities on natural language input in the zero-shot or few-shot learning setting~\cite{wei2022emergent} making them ideal for harm mitigation. We find that, as opposed to traditional harm classifiers such as Perspective API\footnote{\scriptsize{\url{https://www.perspectiveapi.com/}.}}, LLMs can excel at mitigating exposure to harm in recommended content sequences simply by pairwise comparisons and re-ranking, most likely due to better utilization of their context window. The benefits of pairwise re-ranking have been explored in past work \cite{qin2023large} and we adopt this re-ranking strategy using LLMs to mitigate exposure to harmful content on platforms. This approach is visually depicted in Figure \ref{fig:overview}.

We propose two novel metrics to dynamically analyze exposure to harmful content in recommendation sequences and experiment with three LLM preference re-ranking methods (zero-shot, few-shot, in-context learning based~\cite{dong2022survey}). Through extensive experiments on three harm datasets and three LLM architectures, we find that our approach outperforms state-of-the-art content moderation baselines, such as Perspective API and OpenAI's Moderation API.\footnote{\scriptsize{\url{https://platform.openai.com/docs/guides/moderation}.}} In summary, we make the following contributions:
\begin{itemize}
\item We propose a novel LLM-based approach that can measure relative harm and mitigate harmful content exposure on platforms. Our approach is able to generalize to various kinds of harms without explicit training. 
\item We examine the performance of our approach under three settings (zero-shot, zero-shot with prompt engineering and few-shot) by varying the amount of information provided to the LLM, and find that it outperforms industry-grade classifiers even in the zero-shot setting.
\item Alongside evaluation on traditional metrics, we propose two novel metrics that can measure harm exposure for a given content sequence. Our metrics allow evaluating the quality of re-ranked content in a manner that is agnostic to the amount of harm present wherein.
\end{itemize}

This paper is organized as follows. First, we discuss related work in the areas of harms caused by recommendation systems and efforts towards mitigation to contextualize the novelty of our approach. Then, we present our methodology at a high level and introduce two novel metrics for evaluating re-ranking. Next, we describe our experimental setup and analyze our results. Finally, we discuss the limitations of our work and its implications. 

\section{Related Work}
\subsection{Recommendation-Driven Harms}
Platform recommendation systems are designed to maximize user engagement by tailoring content to individual preferences and interests. However, these systems can inadvertently drive users toward harmful content. This occurs because recommendation algorithms often prioritize content that generates the most interaction, which can include sensational or provocative material~\cite{rathje2021out, yu2024partisanship}. For instance, %platforms like TikTok are criticized for feeding users a narrow range of content that increases platform addiction~\cite{klug2021trick}, while 
Facebook's algorithmic curation is linked to the widespread use of clickbait, which drives users toward low-quality content~\cite{lischka2021clickbait}. 
%In addition, recommendation systems are susceptible to \textit{algorithmic negativity bias} a phenomenon where negative content is promoted disproportionately, which tends to generate higher engagement but also increases the visibility of toxic users~\cite{chavalarias2023can}. 

In addition, digital traces left by each user on platforms reveal information about the user’s emotions \cite{hossain2019emotion}, substance use \cite{kosinski2013private}, or sexual orientation \cite{wang2018deep}. This inferred information could be used to recommend content that exposes users---especially the vulnerable ones---to online harms (e.g, addictive content to users known to use substances, suicidal content to depressed users, misinformation to users interested in herbology) \cite{tiktok21}. Indeed, meta-reviews show that 8\%-10\% of recommendations pose detectable risks to users \cite{hilbert20248} and algorithmic audits detect discriminatory or otherwise harmful biases in YouTube, Instagram, and TikTok algorithms \cite{haroon2023auditing, bandy2021problematic, hilbert2023bigtech}. 

Because recommendation systems are exceptionally good at curating personalized ecosystems, leading to closed loops of content consumption~\cite{rossi2021closed}, what they recommend has implications for what users see~\cite{nyhan2023like}. In turn, content exposure has over-time effects on individuals and, when  harmful, can have severe consequences on mental health and may foment addictions, violence, or even lead to death from dangerous challenges~\cite{haidt2022adolescent}. 

%\noindent \textbf{Harm Mitigation. }
\subsection{Harm Mitigation}
\looseness-1Extensive work studies harm mitigation interventions on social media platforms. For example, \citet{bhargava2019gobo} developed a tool that allows users to exert more control over their social media feeds by enabling them to consolidate and tailor content from multiple platforms. Similarly, \citet{kovacs2018rotating} empower users to manage their social media engagement goals via rotating time or site blockers. More germane to minimizing harmful content, some studies explore ranking-based interventions \cite{celis2019controlling, ovadya2023bridging}, such as using LLMs for curating societally beneficial recommendations. In addition, machine learning models, particularly those leveraging deep learning, are often adopted to identify patterns indicative of harmful content, such as clickbait~\cite{ghosh2022detection}, hate speech~\cite{mossie2020vulnerable, del2017hate} and misinformation~\cite{wu2019misinformation, shu2017fake}. Recent work additionally applies LLMs to the harm detection problem ~\cite{liu2024detect, bonagiri2025towards, ernst2024identifying} and also utilizes machine learning toxicity detection algorithms within a browser extension to automatically hide toxic text content on users' feed over-time \cite{beknazar2022toxic}.\vspace{1mm}

%Other work has explored ranking based interventions \cite{celis2019controlling, ovadya2023bridging}, including the role LLMs can play in embedding societally beneficial values into social media recommendation models in simulated settings (mock social media app) \cite{jia2023embedding}. 

%No need to have the title Novelty - AC\
% \subsection{Novelty \& Motivation}
\looseness-1 \noindent\textbf{Remark.} While prior works focus on specific harms, we consider a general, systematic, and overarching taxonomy of harms with data originating from a real social media platform (YouTube), as detailed below. Our approach generalizes to various kinds of harms, as opposed to being effective on one particular kind, like hate, misinformation, or clickbait. We leverage LLMs and their inherent reasoning abilities to reduce harmful content exposure in recommended content. Additionally, because we do not require explicit labeling and training, our approach is robust against \textit{concept drift} \cite{quinonero2022dataset}. Finally, instead of absolute harm, we focus on measuring relative harm and propose novel metrics to quantify the ordering of content for minimal exposure to harm.

\section{Mitigating Harmful Content Exposure Using LLMs and Re-ranking}%\vspace{-0.75mm}

\subsection{Problem Formulation}
The problem we seek to solve is as follows: \vspace{2mm}

\noindent \fbox{%
    \parbox{0.455\textwidth}{\textit{Given a recommendation sequence (e.g. homepage videos on YouTube), re-rank the content so that harmful content appears at the end of the sequence in a zero-shot or few-shot setting (i.e. limited annotations are required)}.}} \vspace{2mm}

\noindent The motivation for downranking comes from past work that has shown that users are less likely to interact with (and be exposed to) content that appears at lower ranks in recommendations~\cite{yu2023nudging, glick2014does}. Downranking, as opposed to outright suppressing or deletion, minimizes exposure to harmful content while preserving transparency, diverse viewpoints and preferences, and freedom of expression.
%Moreover, since our re-ranking based solution to this problem is general and can also be applied to problem domains other than harm mitigation, we frame it more generally as \textit{LLM Preference Re-ranking}. Note that for our use case: \textit{preferred content} $\leftrightarrow$ \textit{non-harmful content} and vice versa.
%
\looseness-1 Formally, let $X = \{x_i\}_{i=1}^n$ be a sequence of $n$ content instances, out of which $p$ are non-harmful, and the remaining $n-p$ are harmful. 
Let $\rho \colon X \rightarrow \{0,1\}$ be a binary decision function that maps every $x$ to a label based on whether it is harmful or not. Our goal is to use an LLM $\mathcal{L}$ to transform $X$ into another sequence $X^*$ such that it minimizes exposure to harmful content.

\subsection{Preferential Pairwise Ranking}
We present our proposed solution to the harm mitigation problem as Algorithm \ref{alg:reorder_content}. Our proposed approach consists of a \textit{pairwise ranking component} that seeks to downrank content if deemed harmful by the LLM $\mathcal{L}$. Specifically, we present $\mathcal{L}$ with pairs of content instances and query it to determine which is the relatively harmful one. Certain preference constraints ($\mathcal{C}$) determine the exact prompt used to query the LLM.
To re-rank content based on relative harm, we adopt the approach from \citet{qin2023large} but modify their methodology and scoring function. For a given content instance $x$, their scoring function increments the score by 1 for every content instance deemed less \textit{relevant} than $x$, and by $0.5$ for all other content instances. However, applying the scoring function as is will also compare non-harmful content with one another and incorrectly result in scores being higher. This can then lead to them being unfairly downranked. 
To address this challenge, in the pairwise ranking process, we allow the LLM $\mathcal{L}$ to decide if both instances are \textit{non-harmful}, in which case there is no increment to the score. Given a content sequence $X$, we enumerate all possible pairs and compute the score for every content instance. We then re-order the content based on this score obtained.\vspace{4mm}

\begin{algorithm}%[H]
\label{algo:approach}
\fontsize{10}{10}\selectfont %9.5
\caption{Harm-Based Re-ranking Using LLMs}\label{alg:reorder_content}
\begin{algorithmic}[1]
\State \textbf{Input:} Sequence \( {X} = \{x_i\}_{i=1}^n \), LLM $\mathcal{L}$, Preference Constraints $\mathcal{C}$
\State \textbf{Output:} Re-ranked Sequence $X^*$

\State \textbf{initialize} \( {score}[x_i] \leftarrow 0 \) for each \( x_i \in {X} \)
\For{each pair $ (x_i, x_j) \in X \times X, \, i \neq j$ }
    \State \textbf{query} $\mathcal{L}$ with $\mathcal{C}$ for pairwise preference: $(x_i, x_j)$ and $(x_j, x_i)$
    \If{\( x_i \) harmful}
        \State \( {score}[x_i] \leftarrow {score}[x_i] + 1 \)
    \ElsIf{\( x_j \) harmful}
        \State \( {score}[x_j] \leftarrow {score}[x_j] + 1 \)
    \Else
        \State \textbf{continue}
    \EndIf
\EndFor
\State \textbf{sort} \( {X} \) using \( {score} \) (ascending) to \textbf{obtain} $X^*$
\State \Return sorted sequence \( {X^*} \)
\end{algorithmic}
\end{algorithm}%\vspace{-3mm}

\subsection{Specifying Preference Constraints $\mathcal{C}$}
\label{sec:constraints}
As described in Algorithm~\ref{alg:reorder_content}, the pairwise ranking via LLMs also requires natural language preference constraints $\mathcal{C}$ as part of the prompt to effectively downrank harmful content. We employ three approaches in specifying the preference constraints for the LLM in the context window/prompt. We mention the approaches here, and offer the detailed prompts in Appendix~\ref{app:prompts_icl}.\vspace{1mm}

\noindent \textbf{Zero-Shot:} In the \textit{zero-shot} setting, $\mathcal{C}$ asks to identify which of the two provided content instances is harmful, without explicitly specifying a definition of \textit{harm}. In this approach, we utilize the LLM's inherent understanding of harm learnt during pretraining.\vspace{1mm}

\noindent \textbf{Zero-Shot + Prompt Engineering:} We build upon the zero-shot approach by including a definition for harm in $\mathcal{C}$. We first define explicitly what we consider harmful and characteristics of harmful content, and then query $\mathcal{L}$ to identify which of the two content instances is harmful. \vspace{1mm} %Note that while we provide a definition of harm, we do not provide any explicit examples of what constitutes harm.

\noindent \textbf{Few-Shot ICL:} We now provide representative instances of harmful content in $\mathcal{C}$, and query $\mathcal{L}$ to reason which of the two content instances is harmful based on this information. This approach is known as In-Context Learning (ICL) \cite{dong2022survey, brown2020language, askari2025unraveling}. %We also experimented with including the definition of harm along with the labeled examples, but we did not see meaningful performance gains.
%More details on the exemplar selection can be found in Appendix~\ref{app:icl}.

\subsection{In-Context Learning (ICL)}
In-Context Learning~(ICL) \cite{dong2022survey} relies on exemplars that the model is exposed to in order to learn certain features or characteristics. Therefore, the performance of our re-ranking approach will greatly depend on the chosen exemplars.
A naive approach of ICL would be to randomly sample from harmful content and provide these random exemplars to the model; however this may lead to bias towards or against certain kinds of harm. 
%A naive approach of ICL would be to randomly sample from harmful content and provide these random exemplars to the model. However, there are several different kinds of harm; random sampling may over- or under-sample these. As a result, the model will be biased towards certain categories of harm. For example, if we do not sample sexually harmful content, we may lead the model to believe that those videos are not harmful. 
%
To address this, we curate a set of exemplars that is representative of the harm in our dataset. Inspired by a popular coverage-based BertScore ICL selection approach~\cite{gupta2023coverage}, we make some modifications for undertaking ICL in the re-ranking setting. We first use a pretrained RoBERTA model to project text content into an embedding space. We cluster the harm samples, and then choose the most representative sample from each cluster (the one that is closest to the centroid of the cluster). These representative samples form the exemplars we provide to the model.

\subsection{Evaluating Re-rankings}
While we use existing metrics for evaluating rankings~\cite{sebe2000toward} such as Precision@K \cite{shani2011evaluating}, we also design new metrics that focus on the \textit{relevance} of ranked content to a query and help account for harmful content that users could be exposed to in the sequence. 
Thus, we propose two novel metrics to assess preference in re-ranked content sequences. Note that both metrics are bounded between $[0,1]$ and higher values are better (less harmful content).\vspace{1mm}

\noindent \textbf{Per-Pref-$k$:}
The Per-Pref-$k$ (PP$k$) metric represents the fraction of the content sequence set $X$ that would need to be consumed to reach the $k$-th harmful content instance. PP$k$ assesses how much of the sequence a user needs to consume before encountering a certain amount of harmful content. This metric is instrumental in understanding the depth of user engagement required to reach less desirable content, thus indirectly measuring the buffer of harmless content. Higher values indicate that a user can view more content before encountering a specified number of harmful instances, reflecting better performance of the moderation system. Formally, we define PP$k$ as:
\begin{equation}
    \text{PP}k = \frac{\min \left\{ m \mid \sum_{i=1}^{m} \rho(x_i) = k \right\}}{n}
\end{equation}

% \begin{definition}
%     \textit{The Top-Pref-$k$ (TP$k$) metric represents the fraction of preferred/non-harmful content the user is exposed to when they consume a $k$-length content sequence $X$.}
% \end{definition} 

% \begin{definition}
%     \textit{The Per-Pref-$k$ (PP$k$) metric represents the fraction of the content sequence set $X$ that would need to be consumed to reach the $k$-th harmful/non-preferred content instance.}
% \end{definition} 

\noindent \textbf{Exponentially Weighted Normalization:}
The Exponentially Weighted Normalization ({EWN}) metric provides an analytical measurement of non-harmful rankings by assigning exponentially decaying weights to ranks and then normalizing values to lie between $[0,1]$. EWN($X$) measures the goodness of the ranking in the sequence $X$ by comparing it to the best and worst ranking possible. A value of $1$ indicates that the sequence is in the optimal order and no better order can minimize exposure to harm. On the other hand, a value of $0$ indicates that the sequence is in the worst possible ranking order. Following the notation described earlier, the EWN can be defined as follows:
\begin{equation}
    \text{EWN} = \frac{\sum_{i=1}^n \{2^{-i}\cdot (1-\rho(x_i))\} -  (2^{-p} - 2^{-n})}{(1-2^{p-n}) \cdot (1-2^{-p})}.
\end{equation}
%EWN values allow for an apples-to-apples comparison of sequence rankings, even if the sequences are drawn from different distributions. 
Owing to space limitations, we provide the complete derivation for EWN in the appendix.\\

\noindent \textbf{Remark.} In addition to these two novel metrics, we also use the \textbf{Precision@K} metric~\cite{shani2011evaluating}, popular in recommendation systems. In line with our naming convention, we call this metric as the \textbf{Top-Pref-$k$ (TP-$k$)}, which represents the fraction of non-harmful content the user encounters when they consume a $k$-length content sequence.

% \begin{definition}
% \textit{The Exponentially Weighted Normalization ({EWN}) metric provides an analytical measurement of preferred/non-harmful rankings by assigning exponentially decaying weights to ranks and then normalizing values to lie between $[0,1]$. }
% \end{definition}
% \noindent Owing to space limitations, we provide the formulation and derivation for EWN in Appendix~\ref{appendix:ewn}.

%\raj{$\rho, p , n$ is not defined and the equation is very confusing to look at without much context. Cannot have eq without notation defined. Would suggest moving to Appendix. It might take up more space to define everything.}

\section{Experimental Setup}
%\subsection{YouTube Harm Dataset and Simulating Recommendation Content Sequences}
\subsection{Datasets}
\looseness-1 We employ a curated dataset of YouTube videos~($9,832$ harmful, $2,679$ harmless), which were labeled for six categories of harm: information, hate and harassment, addictive, clickbait, sexual, and physical harms. The details on the harm categories, data collection, labeling, reliability, and other aspects of this dataset are provided in ~\cite{jo2024harmful, jo2025dataset}. Note that we primarily utilize this dataset for the majority of our ablation experiments because of the diversity in harm categories. We use the video descriptions as input to the LLM. To demonstrate the generalizability of our approach, we also evaluate it on two singular-harm category datasets; the \textit{Jigsaw Toxicity} Classification Dataset~\cite{jigsaw_unintended_bias_2019}, which contains comments from the Civil Comments platform labeled for toxicity at Jigsaw, and the \textit{Measuring Hate Speech} dataset~\cite{ucberkeley_dlab_measuring_hate_speech} by UC Berkeley's D-Lab, that contains annotated social media posts specifically labeled for hate speech. Additional details on each dataset are provided in Appendix~\ref{appendix:data}.

%\noindent \textbf{Content Sequences. }
\subsection{Content Sequences}
\looseness-1 Note that our proposed approach operates on content sequences rather than individual instances. We use the datasets described to simulate sequences of content that a user would be exposed to. We sample uniformly at random without replacement from our data to generate a sequence of $n=20$ content instances (textual data; either titles and transcripts from YouTube videos, comments, or social media posts, depending on the dataset), and generate $m=100$ such sequences to form a sequence dataset. To study the effect of the \textit{amount} of harm, we generate $5$ such datasets by the varying the fraction of harmful content from $10\%$ to $50\%$ for the YouTube dataset. For the other datasets, we fix the fraction of harmful content to $30\%$ so as to reflect the typical harm ratio observed in the wild, as determined by user surveys ~\cite{mostafavi2020young}.

\subsection{Baselines}
We leverage two state-of-the-art harm classification models as benchmarks to compare the performance of our LLM re-ranking approach. %Specifically, we use these models to obtain a score for each content instance, and use these scores to re-rank the content such that the most harmful ones are downranked. 
The Perspective API~\footnote{\scriptsize{\url{https://perspectiveapi.com/}}} is a tool developed by Google/Jigsaw to improve conversations online by detecting toxicity of comments. We use the toxicity score the model returns to rank the content. 
Similar to Perspective, the OpenAI Moderation API\footnote{\scriptsize{\url{https://platform.openai.com/docs/guides/moderation}}} can be used to check whether a text is harmful or not. It provides scores for a variety of harm categories, ranging from hate to sexual or violence; we extract the scores for each category and use the highest score as a proxy for harm.

\subsection{Implementation Details}
We implement our approach using mainly OpenAI GPT-3.5 Turbo as the underlying LLM. This choice was made based on the available models, API costs, and rate limitations at the time of the initial writing. Additionally, we ran experiments using two open-source models: Llama2-13B and Mistral-7B-Instruct-v0.2. The exact prompts we use for each of our approaches are listed in Appendix~\ref{appendix:llmprompts}.

\subsection{Evaluation}
We use three metrics to evaluate the effectiveness of re-ranking: TP-$k$, PP-$k$ and the EWN as described. Given that our sequences are of $20$ videos, we limit our analyses to TP$5$ and TP$10$, as they represent the first two quartiles of videos watched. Similarly, for PP-$k$, we focus our analysis on PP$1-3$, as they represent the amount of content needed to be watched to reach up to at most the third harmful video, which is a practical limit considering the length of our sequences. The PP$1$, PP$2$ and PP$3$ values represent the fraction of the sequence that can be consumed before encountering the first, second and third harmful video(s), respectively.

\section{Results}
\begin{figure*}
\centering
\includegraphics[width=0.88\textwidth]{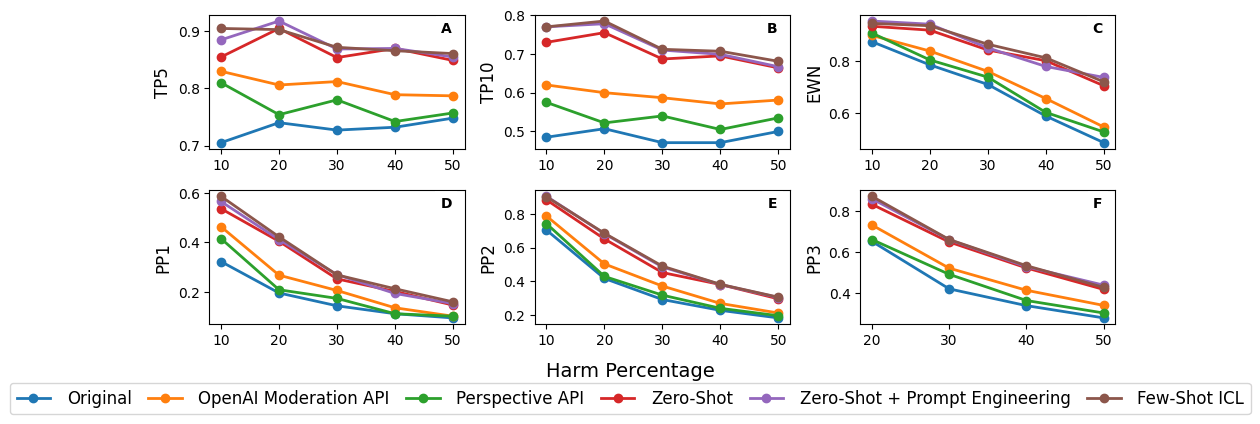}\vspace{-3mm}
\caption{Performance of our proposed method with different preference constraint strategies on varying ratios of harmful content. Higher values indicate better alignment with less harmful content exposure. Our approaches outperform all baselines by a wide margin for the TP$k$ (\textbf{A, B}), PP$k$ (\textbf{D-F}) and EWN (\textbf{C}) metrics. Note that as the harm ratio increases in content sequences, exposure to harmful content increases as well. %\\\raj{I think you can remove "Prompting Approach" from the legend, will save us space and it is evident anyway.}
}
\label{fig:tp5}\vspace{-4mm}
\end{figure*}

\subsection{Effectiveness of LLM Preference Re-ranking}% in Mitigating Harm Exposure}
\looseness-1 Figure~\ref{fig:tp5} depicts the effectiveness of our approaches with respect to the baselines, in terms of our metrics. 
In general, our approach outperforms both the baselines across varying harm percentages, demonstrating the improvement preference re-ranking drives over simple classification. We find that LLM-based approaches outperform standard classification approaches even in zero-shot settings, and even more so in few-shot settings. For example, $TP5$~(Figure~\ref{fig:tp5}A) indicates the average proportion of non-harmful videos in the first five videos in the sequence. This proportion is $70.5\%$ and $74.8\%$ in the $10\%$ and $50\%$ harm settings initially. The OpenAI Moderation API is able to drive these up to $83\%$ and $78.7\%$. The preference re-ranking approach in the Few-Shot ICL setting is able to increase the non-harmful videos to $90.5\%$ and $86.1\%$ respectively.
Mean TP$5$ and TP$10$ values are shown in Tables~\ref{table:tp5} and~\ref{table:tp10} respectively.
Across both tables, the more advanced configurations (Zero-Shot with PE, Few-Shot ICL) consistently show higher TP\textit{k} values across all harmful content percentages, demonstrating the advantage of leveraging sophisticated AI techniques. Notably, the effectiveness of the OpenAI Moderation and Perspective approaches varies, often underperforming compared to the LLM-based approaches, despite the massive amounts of \textit{task-specific} data they are trained on.

\subsection{Effect of Harm Ratio}
%We observe that harm mitigation becomes increasingly challenging as the amount of harmful content increases. Note that the EWN is a metric agnostic of the amount of harm (see Appendix~\ref{appendix:ewn} for explanation); thus, decreasing values of EWN~(Figure~\ref{fig:tp5}E) indicate the reduced effectiveness of \textit{all} approaches, showing that harm mitigation is harder with higher amounts of harmful content. 
%
%The EWN metric measures the effectiveness of re-ranking approaches by applying exponentially decaying weights to content ranks, with the results normalized to a scale from 0 to 1. Higher EWN values indicate better performance in maintaining a sequence of non-harmful content. 
In Table~\ref{table:ewn}, we present EWN values for each constraint specification approach across various harm percentages.
We see that the difference in EWN values between the original and re-ranked sequences grows as the percentage of harmful content increases. For instance, at 10\% harmful content, the gap between the Original and Zero-Shot with PE is about 0.079 (9.1\% relative increase), while at 50\% harmful content, the gap extends to 0.250 (51.3\% relative increase).
Additionally, as the percentage of harm increases, OpenAI Moderation and Perspective, show large decreases in their EWN values~(39.1\% and 41.8\% respectively). Zero-Shot with PE and Few-Shot ICL, on the other hand, exhibit the smallest decreases in their EWN scores as the percentage of harmful content increases. Both configurations manage to minimize the performance drop to around 23\%.
Given that EWN is agnostic to the relative harm ratio (see Appendix~\ref{appendix:ewn}), this analysis shows that as the operational environment becomes more challenging due to higher concentrations of harmful content, the advantage of deploying advanced AI-based content moderation systems becomes increasingly substantial.

\subsection{Varying Exemplars for Few-Shot ICL}
\begin{figure}[t]
\centering
\includegraphics[width=0.28\textwidth]{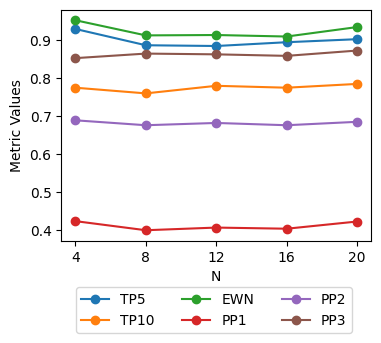}\vspace{-3mm}
\caption{Performance of ICL with varying number of exemplars, as measured by our defined metrics. Harm mitigation effectiveness does not meaningfully improve by increasing the number of exemplars.
}
\label{fig:iclv}\vspace{-3mm}
\end{figure}

%\looseness-1
Because the performance of ICL will depend on exemplars, we experimented with various values for the number of exemplars $N$.
We modify the few-shot ICL approach and implement it in five versions by setting the value of $N$ as one of $4, 8, 12, 16$ and $20$. We perform K-Means clustering on the harm samples (setting $K=N$) and choose a representative sample from each cluster. We conduct our analysis in the $30\%$ setting ~\cite{mostafavi2020young}.
Table~\ref{table:icl_variations} and Figure~\ref{fig:iclv} show the performance of ICL-based prompting with varying number of exemplars. Surprisingly, the performance does not necessarily improve with increasing number of exemplars; we see that EWN is the best for $N=4$ and decreases with increasing $N$. This is likely due to overfitting the LLM and the bias introduced because of the exemplars. Nevertheless, ICL still remains a better approach as compared to zero-shot or zero-shot with prompt engineering.

% \begin{table}[t]
% \centering
% \footnotesize
% \caption{Metrics for variations in the number of exemplars provided as part of ICL.}
% \resizebox{0.49\textwidth}{!}{
% \begin{tabular}{ccccccc}
% \hline 
% {N} & {TP5} & {TP10} & {EWN} & {PP1} & {PP2} & {PP3}\\ \hline \hline
% %Original & 0.740 & 0.507 & 0.785 & 0.195 & 0.417 & 0.653\\
% 4 & 0.930 & 0.775 & 0.953 & 0.423 & 0.689 & 0.853 \\ 
% 8 & 0.887 & 0.760 & 0.913 &0.399 & 0.676 & 0.865  \\ 
% 12 & 0.885 & 0.780 & 0.914 & 0.406 & 0.682 & 0.863 \\
% 16 & 0.895 & 0.775 & 0.910 & 0.403 & 0.676 & 0.859  \\ 
% 20 & 0.903 & 0.785 & 0.935 & 0.422 & 0.685& 0.873 \\ \hline \hline
% \end{tabular}
% }
% \label{table:icl_variations}
% \end{table}
% %We find that our approach works best with $N=4$ exemplars, and that more samples may lead to overfitting and bias (please refer to Appendix~\ref{app:icl} for detailed results).

\subsection{Experiments across Datasets}

The results described so far are based on the YouTube dataset~\cite{jo2025dataset} because of the comprehensive and diverse nature of harms it contains. However, we also find that our LLM-based re-ranking approach performs well on other, more targeted datasets. Table~\ref{table:dlab} shows that all three configurations (zero-shot, zero-shot with prompt engineering and few-shot) outperform the baselines (albeit by a small margin) on the hate speech dataset as evidenced by the EWN values. Notably, the Zero-Shot approach slightly outperforms the other configurations with a perfect TP10 score of 1.000 and slightly higher PP1 (0.697), PP2 (0.777), and PP3 (0.839) values, indicating a marginally better delay in encountering harmful content. The inclusion of Prompt Engineering and Few-Shot learning yields EWN scores very close to the perfect mark (0.99992 and 0.99991, respectively), though they slightly trail the pure Zero-Shot method. On the Jigsaw dataset (Table~\ref{table:jigsaw}), our approach achieves a comparable performance to the baselines. Note that the Jigsaw dataset contains content labeled for toxicity, on which the Perspective API is likely trained (as both are released by Jigsaw). Perspective API attains uncharacteristically high performance on this dataset, potentially indicating \textit{test set leakage} in this experiment.

\subsection{Experiments across LLMs}

The results above have been based on the performance of GPT-3.5 as the primary LLM of choice.
However, there are growing concerns about the lack of transparency and data-sharing practices in closed-source LLMs \cite{balloccu2024leak}, which might deter developers from utilizing them. Further, the costs associated with querying the LLM at scale can also compound their use.
In this section, we thus consider alternative open-source LLMs and re-run experiments with 30\% harm, the typical harm ratio observed on platforms \cite{mostafavi2020young}. We repeat the experiments conducted on the YouTube dataset using two LLMs, namely \textit{Mistral-7B-Instruct-v0.2}\footnote{\scriptsize{\url{https://huggingface.co/mistralai/Mistral-7B-Instruct-v0.2}}} and \textit{Llama2-13B}\footnote{\scriptsize{\url{https://huggingface.co/meta-llama/Llama-2-13b-chat-hf}}}, both running locally on a server equipped with a NVIDIA RTX A6000 GPU with 256GB of RAM.

%%

\begin{figure}[t]
    \centering
    \includegraphics[width=0.8\linewidth]{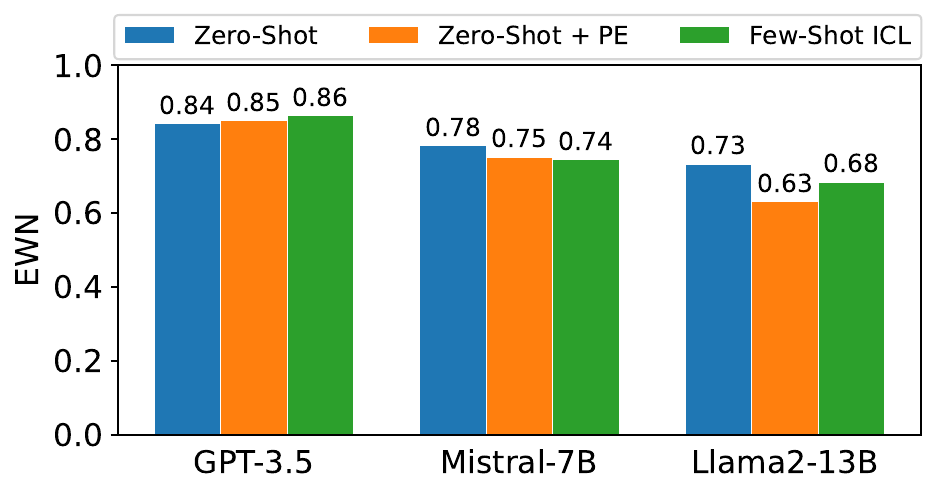}\vspace{-1mm}
    \caption{EWN values for Llama2, Mistral, and GPT-3.5-Turbo.}
    \label{fig:ewn_bar_chart}\vspace{-3mm}
\end{figure}

% \looseness-1 Tables \ref{table:metrics_mistral} and \ref{table:metrics_llama} which detail our various metric values for the Mistral and Llama2 respectively, show that both models perform better than the baselines and slightly worse than GPT-3.5 for all metrics. In addition, Mistral performs better than Llama2 for all metrics.
Table \ref{table:metrics_mistral} shows the various metric values for the YouTube dataset using Mistral.
We see that it too outperforms the baselines significantly and only has slightly lower performance compared to GPT-3.5-Turbo.
In contrast, Table \ref{table:metrics_llama} that Llama2 under-performs across metrics compared to the OpenAI Moderation baseline.
Comparing the models' performances with GPT-3.5 directly, Figure \ref{fig:ewn_bar_chart} shows the EWN values for all three models and each of the three learning strategies we propose.
As is evident, GPT-3.5 outperforms both Mistral and Llama2 (likely) due to its larger parameter size.
However, the EWN values of Mistral trail GPT-3.5 by an average of nearly 10\%, a minimal performance trade-off that make Mistral a viable second option for our approach. Also note that Mistral has only 7B parameters, making it extremely lightweight to run locally and ingest data at scale.
In contrast, Llama2 exhibits lower performance than the other two LLMs, a result also consistent with prior work which has found that Mistral outperforms Llama2 on most benchmark datasets \cite{jiang2023mistral7b}.
%
% Its failure to outperform the baselines, larger parameter size (i.e., higher compute costs) than Mistral, and worse performance than Mistral make it a disadvantageous choice of LLM for re-ranking harms.
%As such, we believe given Llama's larger parameter size (i.e. more compute overhead) and lower performance, Mistral is a better open-source alternative for social media content re-ranking to mitigate harm.

\section{Discussion}

\looseness-1In this work, we propose methods that leverage LLMs to circumvent two challenges in online harm mitigation: the large scale of annotation required and the dynamic nature of harm.\vspace{1mm}

\noindent \textbf{Utility of LLMs for Harm Mitigation.}
\looseness-1 Our experiments demonstrate that our LLM-based re-ranking approaches significantly outperform traditional/proprietary content moderation methods in reducing harmful content exposure on social media platforms. As detailed above, both the Zero-Shot and Few-Shot In-Context Learning (ICL) configurations provide a notable improvement over industry-standard harm classifiers. Because our approach shows promising results even in zero-shot settings, LLMs can be used off-the-shelf with minimal effort and without necessitating significant re-configuration or fine-tuning. Customizing LLM prompts can lead to even better results; for example, the Zero-Shot + Prompt Engineering configuration, where harm is explicitly defined in the prompt, consistently performs better than the simpler Zero-Shot approach, indicating that our automated prompt adjustments can significantly influence desirable performance outcomes.\vspace{1mm}

\noindent \textbf{Robustness to Concept Drift.}
\looseness-1 Despite the differences in content types and the diversity of harm categories, our LLM approach consistently outperforms traditional methods, suggesting that the method is versatile and can be applied across diverse social media platforms and content/harm moderation challenges. Due to extensive pre-training, LLMs can generalize across various types of harm without needing explicit fine-tuning or labeled examples for each new harm type. Consequently, our approach excels at identifying and mitigating a wide range of harmful content, whether it involves hate speech, clickbait, hate and harrassment, or addictive material, among other categories. While LLMs are still restricted temporally by training data, they can adapt to novel scenarios without requiring continuous retraining, and generalize better than traditional supervised learning methods.\vspace{1mm}

\noindent \textbf{Impact of Harm Ratio on Re-Ranking Effectiveness.}
\looseness-1 Our results show that the efficacy of re-ranking is closely related to the ratio of harmful content within the dataset. Specifically, the Exponentially Weighted Normalization (EWN) metric reveals that as the percentage of harmful content increases, downranking harmful content becomes increasingly more challenging. The advantage of our approach becomes even more pronounced in high-volume online harm scenarios where baselines suffer from significant performance drops, but our LLM-based methods exhibit minimal performance degradation.\vspace{1mm}

 \noindent \textbf{Versatility Across Models.}
 Our experiments across various LLMs show that the performance of our approach will indeed depend on the choice of LLMs. While using commercial LLMs can be expensive, we show that open-source LLMs (such as Mistral) show comparable performance in harm mitigation.
 This means that our approach can be implemented with open-source models that will not result in excessive overhead due to API costs. Furthermore, satisfactory performance using open-source LLMs also ensures that our method can be utilized by developers concerned with data sharing/privacy issues relating to closed-source LLMs. We defer the study of other LLMs in the context of our approach to future work. \vspace{2mm}

 \noindent \textbf{Implications for Preference-Based Ranking.}
 \looseness-1 While we focus on re-ranking content to minimize exposure to harmful material, the underlying approach is general and can be adapted to optimize for various other applications beyond harm mitigation. For example, a platform might choose to re-rank content to elevate material that encourages civic participation or promotes mental well-being, thereby aligning recommendations with broader social goals. The flexibility of LLMs in understanding and applying different ranking criteria makes this approach highly adaptable and scalable to different contexts, where the optimization target could shift from reducing harm to enhancing specific \textit{positive} outcomes for users.

%  \noindent \textbf{Generalizability}

\section{Conclusion}
In sum, we present a novel approach to mitigating exposure to harmful content on social media platforms by leveraging Large Language Models (LLMs) for re-ranking recommendation sequences. LLM-based re-ranking not only outperforms traditional classifier-based commercial moderation systems, such as the Perspective API and OpenAI Moderation API, but also exhibits robustness across various harm scenarios and content types. By utilizing zero-shot, few-shot, and prompt-engineered strategies, our method effectively generalizes to multiple forms of harm without the need for extensive labeled data, addressing both scalability and dynamicity issues inherent in content moderation. The proposed re-ranking technique shows substantial promise in reducing the likelihood of users encountering harmful content, even as the volume of such content increases in users' social media ecosystem (as is the case for the heavy consumers of misinformation, violence, among other harmful contents). For future work, we aim to explore multi-modal input/models to further augment performance and apply our methods to other novel problem domains where content ranking plays a critical role in ensuring user safety and trust.

\clearpage
\section*{Limitations}
While our proposed approach outperforms existing baselines, it has certain limitations as well. The approach involves queries to LLMs, which may be expensive in terms of time and infrastructure cost when considered at the scale that social media platforms operate (although most popular social media platforms operate with infrastructure that can support such intervention strategies\footnote{\scriptsize{\url{https://www.facebook.com/metaai/}}}). For smaller organizations, these issues can be minimized by lightweight (e.g. quantized) high-performance LLMs running locally (e.g. Mistral). Another limitation is modality; we only consider textual input in our algorithm and content sequences, even though visual information (e.g. video frames) could be incorporated from YouTube videos to improve performance further. We defer the study of multi-modal LLMs for re-ranking to future work. Finally, LLMs themselves can possess robustness issues \cite{askari2025assessing, chhabra2024revisiting} in downstream tasks (such as re-ranking), although these issues are exhibited by proprietary harm classification models as well \cite{achara2025watching}. 

\section*{Ethical Considerations}
Through this work, we aim to demonstrate how LLMs can be used to minimize user exposure to harmful content online. Our work does not conduct research with human subjects directly. We do not release any new data; all data used are public, and the authors state they followed required ethics guidelines in the collection phase. 
Sharing user content with third-party LLMs may introduce privacy challenges, and concerns that the data may be used for training without consent. However, social media companies can run open-source LLMs (which, from our findings, have performance comparable to proprietary ones) without relying on third parties, which would ensure that user content does not leave the company. Finally, while the primary intent of our approach is to mitigate exposure to harmful content, there exists the risk of over-moderation; LLMs, despite their advanced reasoning capabilities, might reflect inherent biases from their training data, potentially leading to the marginalization of minority perspectives or controversial opinions. Transparent moderation guidelines, diverse datasets and a human-in-the-loop moderation mechanisms can help safeguard freedom of speech and enforce effective moderation at the same time.

\bibliography{refs}
\newpage

\appendix
\section*{Appendix}

\section{Additional Details on EWN}\label{appendix:ewn}
\subsection{Derivation}
Let $X = \{x_i\}_{i=1}^n$ be a sequence of $n$ content instances, out of which $p$ are harmless, and the remaining $n-p$ are harmful. 
Let $\rho$ be a decision function such that:
\begin{equation}
\rho(x)=
    \begin{cases}
        1 & \text{if $x$ is non-harmful}\\
        0 & \text{if $x$ is harmful}. 
    \end{cases}
\end{equation}

\noindent We want to design a metric that penalizes harmful content instances shown towards the \textit{beginning} of the sequence. Thus, content instances at numerically lower ranks have a higher importance in the metric. We can assign exponentially decaying weights to the content instances based on their rank. Specifically, the $i^{th}$ content instance will have a weight of $2^{1-i}$. Following this, we can compute a score $S$ for the sequence $X$ as:
\begin{equation}
    S(X) = \sum_{i=1}^{n} 2^{1-i}\cdot \rho(x_i).
\end{equation}

\noindent The value of $S(X)$ indicates the goodness of the ranking in sequence $X$. Lower values indicate more harmful content shown early on. However, it is dependent on the number of content instances in the sequence and the harm ratio; as such, it is not a good metric to compare two sequences which differ in these distributions. Therefore, we want to compute a metric that indicates how good the sequence $X$ is compared to the \textit{best possible ordering} of the same content. 

\noindent \textbf{Best Case:} In the best case, all the $p$ non-harmful content instances appear at the beginning of the sequence, Therefore, the score $S(X)$ is 
\begin{equation}
\label{eq:basic_score}
    S(X) = \sum_{i=1}^{p} 2^{1-i}.
\end{equation}
This is a geometric progression with the first term $a=1$ and common ratio $r=2^{-1}$ having $p$ terms. Therefore we have:
\begin{equation}
\label{eq:score_max}
    S_{max}(X) = 2 \cdot (1-2^{1-p}).
\end{equation}

\noindent \textbf{Worst Case:} In the worst case, the harmful content is towards the beginning, i.e. the $p$ harmless content instances all occur at the end of the sequence. Therefore the videos from $i=1 \ldots n-p$ are all harmful. Then the score $S(X)$ is
\begin{equation}
    S(X) = \sum_{i=n-p+1}^{n} 2^{1-i}.
\end{equation}
This is a geometric progression with the first term $a=2^{p-n}$ and common ratio $r=2^{-1}$ having $p$ terms. Therefore we have:
\begin{equation}
\label{eq:score_min}
    S_{min}(X) = 2^{p-n+1} - 2^{1-n}.
\end{equation}

\noindent \textbf{Scaling:} Now, we using the score obtained from~\ref{eq:basic_score} and the upper and lower bounds obtained from~\ref{eq:score_max} and~\ref{eq:score_min} respectively, we can normalize the score to a $(0,1)$ range as
\begin{equation}
    \text{EWN} = \frac{S(X) - S_{min}(X)}
                       {S_{max}(X) - S_{min}(X)}.
\end{equation}
Which after replacing all terms, results in the final expression for EWN:
\begin{equation}
    \text{EWN} = \frac{\sum_{i=1}^n \{2^{-i}\cdot (1-\rho(x_i))\} -  (2^{-p} - 2^{-n})}{(1-2^{p-n}) \cdot (1-2^{-p})}.
\end{equation}

\subsection{Importance of EWN}
While both TP$k$ and PP$k$ are fairly explainable, they vary significantly with the length of the content and the fraction of non-preferential elements. As a result, they cannot be used to compare two sequences which differ in these distributions. The $EWN$, however, measures the goodness of the ranking in the sequence by normalizing it between the best and worst possible ranking possible. A value of $0$ indicates that the sequence is in the optimal order. On the other hand, a value of 1 indicates that the sequence is in the worst possible ranking order.
$EWN$ values allow for an apples-to-apples comparison of sequence rankings. If there are two sequences $X_1$ and $X_2$, then $EWN(X_1) > EWN(X_2) $ implies that the ordering in $X_1$ is better than $X_2$, and this holds even if the sequences differ in length or the fraction of non-preferential content. Here, \textit{better} means less harmful content shown early on.

\section{Additional Details on the Dataset}
\label{appendix:data}
In this section, we provide some additional information about the datasets used in our experiments.
\subsection{YouTube Harms Dataset}
The Harmful YouTube Dataset~\cite{jo2025dataset} consists of metadata (title, transcript, descriptions) from $19,422$ YouTube videos. These were systemically collected using three approaches (i.e., keyword-based, channel-based, and external dataset integration). The videos were labeled by domain experts and crowdworkers as harmful or belonging to one or more categories of harm. The six harm categories include: information harm, hate and harassment, addictive, clickbait, sexual, and physical harms (developed based on prior work and platform guidelines). The details on the harm categories, data collection, labeling, reliability, and other aspects of this dataset are provided in ~\cite{jo2024harmful, jo2025dataset}. The data collection and annotation was IRB-approved and followed the best practices so as to minimize harm to human subjects.

\subsection{Jigsaw Toxicity Dataset}
The Jigsaw Unintended Bias in Toxicity Classification dataset~\cite{jigsaw_unintended_bias_2019} was developed to address the challenge of unintended bias in machine learning models tasked with detecting toxic content. This dataset, also released as part of a Kaggle competition, contains over $1.8M$ comments from the Civil Comments platform, each labeled for toxicity by human moderators. In addition to a primary binary toxicity label, the dataset includes auxiliary labels indicating whether the comment contains identity-based mentions (e.g., referencing a particular race or gender).

\subsection{DLab Hate Speech Dataset}
The DLab Hate Speech Dataset~\cite{ucberkeley_dlab_measuring_hate_speech} consists of $50,070$ social media comments collected from platforms like YouTube, Reddit, and Twitter. These comments were annotated by $11,143$ crowdworkers, primarily sourced from Amazon Mechanical Turk. Each comment was evaluated across $10$ ordinal labels that capture various aspects of hate speech, including sentiment, disrespect, insult, and dehumanization, among others. Each comment is annotated by several moderators and assigned a \textit{hate speech score}. For our experiments, we aggregated the scores by comment and considered the median score to assign a binary label to each comment.

\section{LLM Prompts}
\label{appendix:llmprompts}
In this section, we describe the prompts we used for each of the approaches presented in Section~\ref{sec:constraints}.
\subsection{Zero Shot}
\textit{Read the following texts, A and B. \\
Text A: <<Text A goes here>>\\
Text B: <<Text B goes here>>\\
Which of these two is more harmful? Respond in the form Response=X, where X is one of A, B or NONE.}

\subsection{Zero-Shot + Prompt Engineering}
\textit{Read the following texts, A and B. \\
Text A: <<Text A goes here>>\\
Text B: <<Text B goes here>>\\
We define harm as content belonging to the following categories: 
\begin{itemize}
    \item Information Harms (e.g., Fake news, Conspiracy theories, Unverified medical treatments, Unproven scientific myths) 
\item Hate and Harassment Harms (e.g., Insult, obscenities, Identity attacks or misrepresentation, Hate speech based on gender, race, religion, disability, sexual orientation)
\item Addictive Harms (e.g Game play, Drug promotion, Gambling videos and gambling promotion) 
\item Clickbaitive Harms (Clickbait titles - using exaggerated words, such as "top 10", "best/most", how to", "hacks", get-rich-quick schemes and financial scams, Gossip about celebrities) 
\item Sexual Harms (e.g., Erotic clips extracted from non-pornographic films, Depiction of sexual acts, such as sex and groping, Depiction of clothed or unclothed genitals, breasts, or buttocks) 
\item Physical Harms (e.g, Self-injury, Eating disorder promotion, Potentially dangerous challenges and pranks, Violent content)
\end{itemize}
Based on this definition of harm, which of the two texts (A or B) is more harmful? Respond in the form Response=X, where X is one of A, B or NONE.}

\subsection{Few-Shot ICL}
\label{app:prompts_icl}
In this approach, our prompt is in the form of a multi-dialogue conversation. The OpenAI API allows us to establish prior knowledge by providing a series of dialogues between the user and the LLM. As fitting exemplars in a single problem can prove challenging, we use a multi-step prompt here, as follows. \\
\textbf{User:} \textit{Below are $N$ examples of content that we consider harmful, separated by ten '=' symbols. \\
Exemplar 1 \\ ========== \\
Exemplar 2 \\ ========== \\
. \\
. \\
. \\
==========\\
Exemplar N}\\
\textbf{System:} \textit{Based on these examples, I understand what constitutes harmful content.}\\
\textbf{User:} \textit{Read the following texts, A and B. \\
Text A: <<Text A goes here>>\\
Text B: <<Text B goes here>>\\
Based on the definition of harm you have learned from the samples provided, which of the two texts (A or B) is more harmful? Respond in the form Response=X, where X is one of A, B or NONE.}

\section{Detailed Results}
In this section, we present our detailed results. These results form the basis of Figure~\ref{fig:tp5}.

\subsection{Top-Pref-k~(\textit{TPk})}
The TP$k$ metric measures user exposure to \textit{preferential content} by calculating the fraction of content deemed preferential in the first $k$ items of the sequence. This metric directly addresses the impact of the \textit{initial content} a user encounters, which is often decisive in shaping user experience and perception. Given that our sequences are of $20$ videos, we limit our analyses to TP$5$ and TP$10$, as they represent the first two quartiles of videos watched. Mean TP$5$ and TP$10$ values are shown in Tables~\ref{table:tp5} and~\ref{table:tp10} respectively.

\begin{table}[H]
\centering
\footnotesize
\caption{TP5 values for constraint specification approaches. Columns represent the percentage of harmful videos in the sequence. \textit{PE} $\rightarrow$ \textit{Prompt Engineering}.}\label{table:tp5}
\resizebox{0.49\textwidth}{!}{%
\begin{tabular}{cccccc}
\hline
{Configuration} & {10} & {20} & {30} & {40} & {50} \\ \hline\hline
Original & 0.705 & 0.740 & 0.727 & 0.732 & 0.748 \\ 
OpenAI Moderation & 0.830 & 0.806 & 0.812 & 0.789 & 0.787 \\ 
Perspective & 0.810 & 0.754 & 0.780 & 0.742 & 0.757 \\ 
Zero-Shot & 0.855 & 0.905 & 0.854 & 0.870 & 0.849 \\ 
Zero-Shot + PE & 0.885 & 0.918 & 0.869 & 0.870 & 0.855 \\ 
Few-Shot ICL & 0.905 & 0.903 & 0.872 & 0.866 & 0.861 \\ \hline\hline
\end{tabular}
}
\end{table}

\begin{table}[H]
\centering
\footnotesize
\caption{TP10 values for constraint specification approaches. Columns represent the percentage of harmful videos in the sequence. \textit{PE} $\rightarrow$ \textit{Prompt Engineering}.}
\label{table:tp10}
\resizebox{0.49\textwidth}{!}{%
\begin{tabular}{cccccc}
\hline
{Configuration} & {10} & {20} & {30} & {40} & {50} \\ \hline\hline
Original & 0.485 & 0.507 & 0.471 & 0.471 & 0.500 \\ 
OpenAI Moderation & 0.620 & 0.600 & 0.587 & 0.571 & 0.581 \\ 
Perspective & 0.575 & 0.522 & 0.540 & 0.505 & 0.535 \\ 
Zero-Shot & 0.730 & 0.755 & 0.687 & 0.695 & 0.664 \\ 
Zero-Shot + PE & 0.770 & 0.778 & 0.710 & 0.699 & 0.667 \\ 
Few-Shot ICL & 0.770 & 0.785 & 0.712 & 0.707 & 0.681 \\ \hline\hline
\end{tabular}
}
\end{table}

\subsection{Per-Pref-k~(\textit{PPk})}
We focus our analysis on PP$1-3$, as they represent the amount of content needed to be watched to reach up to at most the third harmful video, which we believe is a practical limit considering the length of our sequences. The PP$1$, PP$2$ and PP$3$ values represent the fraction of the sequence that can be consumed before encountering the first, second and third harmful video(s), respectively. Similar to TP\textit{k} values, the PP$1-3$ values also demonstrate the effectiveness of LLM-based approaches over traditional classifier-based methods. \\

\begin{table}[H]
\centering
\footnotesize
\caption{PP1 values for constraint specification approaches. Columns represent the percentage of harmful videos in the sequence. \textit{PE} $\rightarrow$ \textit{Prompt Engineering}.}
\resizebox{0.49\textwidth}{!}{
\begin{tabular}{cccccc}
\hline 
{Configuration} & {10} & {20} & {30} & {40} & {50} \\ \hline \hline
Original & 0.322 & 0.195 & 0.143 & 0.111 & 0.094 \\ 
OpenAI Moderation & 0.463 & 0.267 & 0.205 & 0.135 & 0.101 \\ 
Perspective  & 0.415 & 0.208 & 0.173 & 0.111 & 0.101 \\ 
Zero-Shot & 0.536 & 0.404 & 0.252 & 0.201 & 0.146 \\ 
Zero-Shot + PE & 0.565 & 0.411 & 0.269 & 0.193 & 0.152 \\ 
Few-Shot ICL & 0.586 & 0.422 & 0.268 & 0.212 & 0.159 \\ \hline\hline
\end{tabular}
}
\label{table:pp1}
\end{table}

\begin{table}[H]
\centering
\footnotesize
\caption{PP2 values for constraint specification approaches. Columns represent the percentage of harmful videos in the sequence. \textit{PE} $\rightarrow$ \textit{Prompt Engineering}.}
\resizebox{0.49\textwidth}{!}{
\begin{tabular}{cccccc}
\hline 
{Configuration} & {10} & {20} & {30} & {40} & {50} \\ \hline \hline
Original & 0.704 & 0.417 & 0.292 & 0.228 & 0.183 \\ 
OpenAI Moderation & 0.787 & 0.503 & 0.372 & 0.269 & 0.212 \\ 
Perspective & 0.742 & 0.429 & 0.318 & 0.240 & 0.196 \\
Zero-Shot & 0.883 & 0.651 & 0.452 & 0.381 & 0.296 \\ 
Zero-Shot + PE & 0.904 & 0.682 & 0.485 & 0.381 & 0.304 \\ 
Few-Shot ICL & 0.901 & 0.685 & 0.490 & 0.382 & 0.306 \\ \hline \hline
\end{tabular}
}
\label{table:pp2}
\end{table}

\begin{table}[H]
\centering
\caption{PP3 values for constraint specification approaches. Columns represent the percentage of harmful videos in the sequence. \textit{PE} $\rightarrow$ \textit{Prompt Engineering}.}
\resizebox{0.49\textwidth}{!}{
\begin{tabular}{cccccc}
\hline 
{Configuration} & {10} & {20} & {30} & {40} & {50} \\ \hline \hline
Original & - & 0.653 & 0.421 & 0.339 & 0.279 \\ 
OpenAI Moderation & - & 0.733 & 0.522 & 0.414 & 0.34 \\ 
Perspective & - & 0.662 & 0.492 & 0.364 & 0.303 \\
Zero-Shot & - & 0.834 & 0.650 & 0.523 & 0.418 \\ 
Zero-Shot + PE & - & 0.861 & 0.661 & 0.529 & 0.439 \\ 
Few-Shot ICL & - & 0.873 & 0.662 & 0.534 & 0.428 \\ \hline \hline
\end{tabular}
}
\label{table:pp3}
\end{table}

\subsection{EWN}
The EWN metric and it's importance have already been described in great detail in Appendix~\ref{appendix:ewn}. The EWN metric measures the effectiveness of these configurations in prioritizing non-harmful content by applying exponentially decaying weights to content ranks, with the results normalized to a scale from 0 to 1. Higher EWN values indicate better performance in maintaining a sequence of preferred content. In Table~\ref{table:ewn}, we present EWN values for each constraint specification approach across various harm percentages.

\begin{table}[H]
\centering
\footnotesize
\caption{EWN values for constraint specification approaches. Columns represent the percentage of harmful videos in the sequence. \textit{PE} $\rightarrow$ \textit{Prompt Engineering}.}
\label{table:ewn}
\resizebox{0.49\textwidth}{!}{%
\begin{tabular}{cccccc}
\hline
{Configuration} & {10} & {20} & {30} & {40} & {50} \\ \hline\hline
Original & 0.874 & 0.785 & 0.710 & 0.589 & 0.487 \\ 
OpenAI Moderation & 0.898 & 0.838 & 0.760 & 0.656 & 0.547 \\ 
Perspective & 0.907 & 0.804 & 0.738 & 0.603 & 0.528 \\ 
Zero-Shot & 0.933 & 0.918 & 0.842 & 0.801 & 0.705 \\ 
Zero-Shot + PE & 0.953 & 0.941 & 0.850 & 0.779 & 0.737 \\ 
Few-Shot ICL & 0.944 & 0.935 & 0.864 & 0.812 & 0.720 \\ \hline\hline
\end{tabular}
}
\end{table}

\subsection{In-Context Learning}
We also study how the performance of preference-based re-ranking varies with the number of exemplars.
Table~\ref{table:icl_variations} and Figure~\ref{fig:iclv} show the performance of ICL-based prompting with varying number of exemplars.
\begin{table}[H]
\centering
\footnotesize
\caption{Metrics for variations in the number of exemplars provided as part of ICL.}
\resizebox{0.49\textwidth}{!}{
\begin{tabular}{ccccccc}
\hline 
{N} & {TP5} & {TP10} & {EWN} & {PP1} & {PP2} & {PP3}\\ \hline \hline
%Original & 0.740 & 0.507 & 0.785 & 0.195 & 0.417 & 0.653\\
4 & 0.930 & 0.775 & 0.953 & 0.423 & 0.689 & 0.853 \\ 
8 & 0.887 & 0.760 & 0.913 &0.399 & 0.676 & 0.865  \\ 
12 & 0.885 & 0.780 & 0.914 & 0.406 & 0.682 & 0.863 \\
16 & 0.895 & 0.775 & 0.910 & 0.403 & 0.676 & 0.859  \\ 
20 & 0.903 & 0.785 & 0.935 & 0.422 & 0.685& 0.873 \\ \hline \hline
\end{tabular}
}
\label{table:icl_variations}
\end{table}

\subsection{Results Across LLM Architectures}
\begin{table}[H]
\centering
\footnotesize
\caption{Results of LLM-based re-ranking using Mistral-7B.}
\label{table:metrics_mistral}
\resizebox{0.49\textwidth}{!}{%
\begin{tabular}{ccccccc}
\hline
{Configuration} & {TP5} & {TP10} & {PP1} & {PP2} & {PP3} &  {EWN} \\ \hline\hline
Original & 0.727 & 0.471 & 0.143 & 0.292 & 0.421 & 0.710 \\
OpenAI Moderation & 0.812 & 0.587 & 0.205 & 0.372 & 0.522 & 0.760 \\
Perspective & 0.780 & 0.540 & 0.173 & 0.318 & 0.492 & 0.738 \\
Zero-Shot & 0.805 & 0.581 & 0.187 & 0.366 & 0.529 & 0.783 \\
Zero-Shot + PE & 0.753 & 0.550 & 0.166 & 0.332 & 0.499 & 0.742 \\
Few-Shot ICL & 0.818 & 0.645 & 0.217 & 0.407 & 0.583 & 0.824 \\ \hline\hline
\end{tabular}%
}
\end{table}

\begin{table}[H]
\centering
\footnotesize
\caption{Results of LLM-based re-ranking using Llama2-13B.}
\label{table:metrics_llama}
\resizebox{0.49\textwidth}{!}{%
\begin{tabular}{ccccccc}
\hline
{Configuration} & {TP5} & {TP10} & {PP1} & {PP2} & {PP3} &  {EWN} \\ \hline\hline
Original & 0.727 & 0.471 & 0.143 & 0.292 & 0.421 & 0.710 \\
OpenAI Moderation & 0.812 & 0.587 & 0.205 & 0.372 & 0.522 & 0.760 \\
Perspective & 0.780 & 0.540 & 0.173 & 0.318 & 0.492 & 0.738 \\
Zero-Shot & 0.778 & 0.545 & 0.178 & 0.324 & 0.489 & 0.742 \\
Zero-Shot + PE & 0.746 & 0.521 & 0.137 & 0.304 & 0.466 & 0.631 \\
Few-Shot ICL & 0.751 & 0.541 & 0.150 & 0.309 & 0.489 & 0.684 \\ \hline\hline
\end{tabular}%
}
\end{table}

\subsection{Results Across Datasets}
\begin{table}[H]
\centering
\footnotesize
\caption{Results on the Jigsaw toxicity dataset. LLM-based re-ranking improves content sequences and achieves performance comparable to baselines.}
\label{table:jigsaw}
\resizebox{0.49\textwidth}{!}{%
\begin{tabular}{ccccccc}
\hline
{Configuration} & {TP5} & {TP10} & {PP1} & {PP2} & {PP3} &  {EWN} \\ \hline\hline
Original & 0.758 & 0.521 & 0.158 & 0.325 & 0.474 & 0.704 \\
OpenAI Moderation & 0.975 & 0.895 & 0.501 & 0.647 & 0.764 & 0.988 \\
Perspective & 0.995 & 0.980 & 0.671 & 0.781 & 0.842 & 0.990 \\
Zero-Shot & 0.953 & 0.818 & 0.398 & 0.587 & 0.702 & 0.962 \\
Zero-Shot + PE& 0.973 & 0.830 & 0.433 & 0.603 & 0.719 & 0.982 \\
Few-Shot ICL & 0.933 & 0.816 & 0.386 & 0.590 & 0.716 & 0.939 \\ \hline\hline
\end{tabular}%
}
\end{table}

\begin{table}[H]
\centering
\footnotesize
\caption{Results on the DLab hate speech dataset. LLM-based re-ranking improves content sequences and outperforms both baselines.}
\label{table:dlab}
\resizebox{0.49\textwidth}{!}{%
\begin{tabular}{ccccccc}
\hline
{Configuration} & {TP5} & {TP10} & {PP1} & {PP2} & {PP3} &  {EWN} \\ \hline\hline
Original & 0.738 & 0.488 & 0.157 & 0.307 & 0.442 & 0.69800 \\
OpenAI Moderation & 1.000 & 0.995 & 0.681 & 0.766 & 0.827 & 0.99990 \\
Perspective & 0.998 & 0.988 & 0.655 & 0.747 & 0.819 & 0.99470 \\
Zero-Shot & 1.000 & 1.000 & 0.697 & 0.777 & 0.839 & 0.99993 \\
Zero-Shot + PE & 1.000 & 0.998 & 0.695 & 0.773 & 0.837 & 0.99992 \\
Few-Shot ICL & 1.000 & 0.998 & 0.695 & 0.770 & 0.834 & 0.99991 \\ \hline\hline
\end{tabular}%
}
\end{table}

\vspace{5mm}
\section{Code and Reproducibility}
Our code and the data used are available at the following repository: \url{https://github.com/rvoak/harm-ranking-llm/}.

\end{document}